\documentclass[letterpaper, 10 pt, conference]{ieeeconf}

\IEEEoverridecommandlockouts                              % This command is only needed if 
\usepackage{graphics} % for pdf, bitmapped graphics files
\usepackage{epsfig} % for postscript graphics files
\usepackage{mathptmx} % assumes new font selection scheme installed
\usepackage{times} % assumes new font selection scheme installed
\usepackage{amsmath} % assumes amsmath package installed
\usepackage{amssymb}  % assumes amsmath package installed
\usepackage{multirow}
\usepackage{cite}
\usepackage{subcaption}
\usepackage{hyperref}
\usepackage{cleveref}
\usepackage[dvipsnames]{xcolor}
\usepackage{graphicx}
\usepackage{booktabs}
\usepackage{multirow}
\usepackage{pifont}
\usepackage{wrapfig}
\usepackage{caption}
\usepackage{xcolor}
\usepackage{tcolorbox}

\makeatletter
\def\thanks#1{\protected@xdef\@thanks{\@thanks
        \protect\footnotetext{#1}}}
\makeatother

\newcommand{\cmark}{\ding{51}}%
\newcommand{\xmark}{\ding{55}}%
\newcommand{\eg}{{\em e.g.}}%
\newcommand{\ie}{{\em i.e.}}%

\title{\LARGE \bf
FOM-Nav: Frontier-Object Maps for Object Goal Navigation
}

\author{
Thomas Chabal${}^{\dagger\ddagger}$
\and
Shizhe Chen${}^{\dagger\ddagger}$
\and
Jean Ponce${}^{\ddagger*}$
\and
Cordelia Schmid${}^{\dagger\ddagger}$
\thanks{
${}^{\dagger}$ Inria.
${}^{\ddagger}$ Department of Computer Science, \'Ecole normale supérieure (ENS-PSL, CNRS, Inria).
${}^{*}$ Courant Institute of Mathematical Sciences and Center for Data Science, New York University.
{\tt\small \{firstname.lastname\}@inria.fr}
}
}

\begin{document}

\maketitle

\begin{abstract}

This paper addresses the Object Goal Navigation problem, where a robot must efficiently find a target object in an unknown environment. 
Existing implicit memory-based methods struggle with long-term memory retention and planning, while explicit map-based approaches lack rich semantic information. 
To address these challenges, we propose \textbf{FOM-Nav}, a modular framework that enhances exploration efficiency through \textbf{F}rontier-\textbf{O}bject \textbf{M}aps and vision-language models.
Our Frontier-Object Maps are built online and jointly encode spatial frontiers and fine-grained object information. 
Using this representation, a vision-language model performs multimodal scene understanding and high-level goal prediction, which is executed by a low-level planner for efficient trajectory generation.
To train FOM-Nav, we automatically construct large-scale navigation datasets from real-world scanned environments.
Extensive experiments validate the effectiveness of our model design and constructed dataset. FOM-Nav achieves state-of-the-art performance on the MP3D and HM3D benchmarks, particularly in navigation efficiency metric SPL, and yields promising results on a real robot.

\end{abstract}

\section{Introduction}
\label{sec:intro}
\vspace{-0.25em}

Autonomous navigation has been a long-standing challenge in robotics~\cite{burgard1998interactivemuseumtour}, dating back to the pioneering work on the robot Shakey~\cite{nillson1984shakey} in the 1960s.
While early work focused on navigating to specific points~\cite{deSouza2002survey,boninFont2008survey} with a preconstructed map~\cite{leonard1991simultaneous,thrun2008slambook}, recent research has progressively shifted towards navigation in unknown environments using textual~\cite{anderson2018vln,batra2020objectnav} or visual~\cite{zhu2017imagegoal} goals, which is an essential capability for enabling mobile manipulation systems~\cite{watkinsvalls2022mobilemanipulation,liu2024okrobot} to perform diverse real-world tasks.

In this work, we focus on the object goal navigation task (ObjectNav)~\cite{batra2020objectnav}, where an agent must navigate to a target object category in an unknown environment using RGB-D observations.
This task requires long-horizon multimodal scene understanding and efficient exploration.
The robot should not only recognize objects within its current field of view but also use previous observations to develop more accurate scene understanding.
Furthermore, effective object-goal exploration demands maintaining long-term memory of visited areas and learning implicit environment priors, \eg, understanding that pillows are more likely to be found in a bedroom or a living room than in the kitchen.

\begin{figure}
    \centering
    \includegraphics[width=\linewidth]{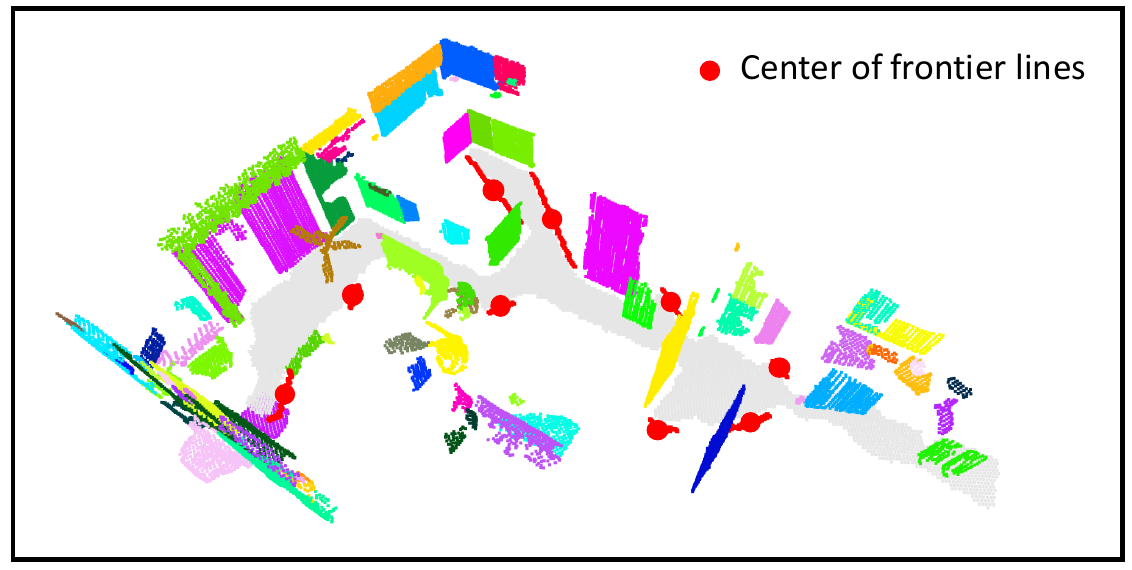}
    \vspace{-1.5em}
    \caption{
    The proposed frontier-object map is a rich representation of objects and frontiers (boundaries of the explored scene), displayed here as colored point clouds and red lines. 
    It encodes geometric, distance and visual/textual information for frontiers and objects.
    }
    \label{fig:teaser}
    \vspace{-2em}
\end{figure}

Prior ObjectNav methods fall into two categories based on their memory representations.
The first approach employs implicit models, typically encoded within hidden states of recurrent neural networks~\cite{ramrakhya2022habitatweb,ramrakhya2023pirlnav,yadav2023ovrlv2} or transformers~\cite{chen2023rim,zhang2024navid,chang2024mobilityvla,zhang2024uninavid}.
While effective for multimodal understanding~\cite{zhou2024navgpt,zhou2024navgpt2,zhang2024navid,chang2024mobilityvla} with large language models (LLMs)~\cite{touvron2023llama,openai2024gpt4} or vision-language models (VLMs)~\cite{liu2023llava,team2023gemini}, they require large-scale training and struggle with long-term spatial memory~\cite{ramrakhya2023pirlnav}, limiting both long-horizon planning and interpretability.
The second approach constructs explicit representations of the environment~\cite{chaplot2020semexp,luo2022stubborn,zhai2023peanut,zhang2024imaginebeforego,yokoyama2024vlfm,huang2023vlmaps}.
Most construct 2D semantic maps~\cite{chaplot2020semexp,luo2022stubborn,zhai2023peanut,zhang2024imaginebeforego} and learn CNN-based policies to predict a long-term navigation waypoint, but these maps cover few object categories, lack semantic relationships across categories (\eg, ``chair" and ``armchair" are similar), and fail to exploit rich visual cues.
VLMaps~\cite{huang2023vlmaps} uses semantic embeddings instead of fixed object labels, yet still loses fine-grained 3D information.
VLFM~\cite{yokoyama2024vlfm} employs a VLM~\cite{li2023blip2} to construct a value map based on image-text similarity, achieving strong zero-shot performance. However, its maps are tailored to a single target object, cannot be reused, and cannot be further improved from robot data.

\begin{figure*}[t]
    \centering
    \includegraphics[width=\linewidth]{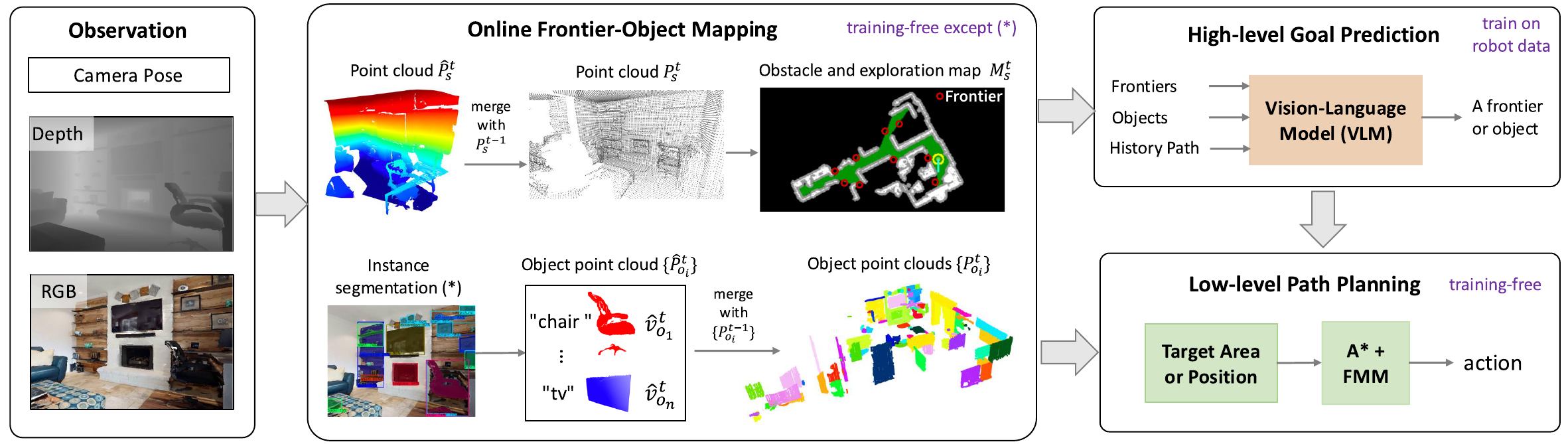}
    \vspace{-1.75em}
    \caption{
    FOM-Nav framework.
    At each step, the agent receives a posed RGB-D image.
    First, we back-project depth into a point cloud to maintain a 3D obstacle map, from which we derive 2D obstacle and exploration maps and frontiers. Simultaneously, we segment objects, and extract and store their geometric, visual and textual features in a 3D object map.
    Then, a VLM processes frontiers, objects, and history path to predict a high-level navigation goal.
    Finally, a low-level path planner (A${}^*$ and fast marching algorithm) plans a path to the goal.
    }
    \label{fig:overview}
    \vspace{-1.5em}
\end{figure*}

In this work, we introduce \textbf{FOM-Nav}, a modular framework with \textbf{F}rontier-\textbf{O}bject \textbf{M}aps for object goal \textbf{Nav}igation (\Cref{fig:overview}).
We propose \emph{online frontier-object mapping} that explicitly maintains navigation memory, including obstacle and exploration maps alongside 3D object point clouds (\Cref{fig:teaser}).
Frontiers~\cite{yamauchi1997frontier,yokoyama2024vlfm} which represent boundary lines between explored and unexplored regions, are extracted from the maps to guide efficient exploration, while object point clouds augmented with visual and text features capture fine-grained semantic representations of the scene. 
Then, the \emph{high-level goal prediction} module leverages a VLM to predict the next navigation goal given the encoded frontiers, objects and past trajectories.
Finally, the \emph{low-level path planning} module converts the high-level goal into an executable trajectory via A$^*$~\cite{hart1968astar} and fast marching algorithms~\cite{sethian1996fastmarching}.
To support model training,
we automatically collect a large-scale robot navigation dataset using real-world scanned environments~\cite{ramakrishnan2021hm3d}.
FOM-Nav achieves state-of-the-art performance on the MP3D and HM3D v1, v2 benchmarks in single-floor settings.
Ablations confirm the contributions of our frontier-object mapping, feature design, and training data.

Our work makes the following key contributions:
\begin{itemize}
    \item We introduce online frontier-object mapping to explicitly capture fine-grained memory with rich spatial-semantic information for object-goal navigation.

    \item We leverage a VLM architecture to predict high-level navigation goals by encoding frontier, object and history path features, and construct an automatic robot navigation dataset to support model learning.

    \item FOM-Nav sets a new state-of-the-art in ObjectNav benchmarks, surpassing prior methods on HM3D v2  with improvements of \textbf{11.8} in success rate and \textbf{14.9} in SPL, and shows promising results in the real world.

\end{itemize}
Our model, code and dataset are publicly available at \url{https://github.com/thomaschabal/fom-nav}.

\section{Related Work}
\label{sec:related_work}

\noindent \textbf{Semantic Mapping for Navigation.}
Semantic maps enhance traditional geometric maps~\cite{thrun2008slambook} by incorporating semantic information such as region types and object names, which are crucial for long-horizon multimodal navigation problems~\cite{raychaudhuri2025survey}.
Building semantic maps requires robot localization, 3D projection, feature extraction and map accumulation, where many methods assume perfect localization for simplicity.
Existing semantic mapping approaches can be categorized by their structural representations, including spatial grids~\cite{chaplot2020semexp,huang2023vlmaps}, topological graphs~\cite{garg2024robohop,chen2022duet}, dense point clouds~\cite{peng2023openscene,zhang20233dawareobjectnav} and hybrid maps~\cite{gu2024conceptgraphs}.
The 2D semantic map~\cite{chaplot2020semexp,ramakrishnan2022poni,luo2022stubborn} is one of the most widely used structures in spatial grids for navigation, but suffers from predefined categories and the lack of rich visual information.
Dense points address these limitations but incur large memory cost.
The hybrid approaches balance these trade-offs, yet their potential for online exploration remains unexplored.
Our work introduces a hybrid representation frontier-object maps to support efficient online object goal navigation.

\noindent \textbf{Object Goal Navigation.}
After decades of focusing on point-based navigation~\cite{nillson1984shakey,thrun2008slambook,deSouza2002survey}, research~\cite{anderson2018evaluation} has shifted towards navigation with goals specified as object categories~\cite{batra2020objectnav}, images~\cite{zhu2017imagegoal} or text instructions~\cite{anderson2018vln}.
This requires multimodal understanding for goal identification and common sense reasoning for efficient navigation.
Current approaches for ObjectNav differ in map representations, exploration strategies and path planning.
One line of work uses implicit representations~\cite{ramrakhya2022habitatweb,yadav2023ovrlv2,ramrakhya2023pirlnav,ye2021auxiliary,majumdar2022zson,chen2023rim} learned from neural networks, which are trained via behavior cloning (BC) or reinforcement learning (RL).
Another approach employs a modular pipeline~\cite{chaplot2020semexp,ramakrishnan2022poni,luo2022stubborn,zhai2023peanut,zhang2024imaginebeforego,zhang20233dawareobjectnav,yokoyama2024vlfm,huang2023vlmaps}: semantic mapping, long-term goal prediction, and path planning.
The goal prediction strategies range from RL with extrinsic rewards~\cite{chaplot2020semexp,zhang20233dawareobjectnav}, to BC with heuristic scores obtained from privileged information~\cite{ramakrishnan2022poni,zhu2022distanceprediction}, scene imagination~\cite{zhai2023peanut,zhang2024imaginebeforego}, and language-grounded value extraction~\cite{yokoyama2024vlfm}.
For path planning, either deterministic planners~\cite{sethian1996fastmarching} or PointNav policies~\cite{wijmans2020ddppo} are utilized.
Our work follows the modular pipeline but introduces an exploration policy powered by vision-language models and proposes frontier-object maps.

\noindent \textbf{Large Language and Vision Models.}
Recently, large language models (LLMs)~\cite{touvron2023llama,openai2024gpt4} have achieved significant success trained with internet-scale data.
They have been extended to vision-language models (VLMs) with inputs of images~\cite{liu2023llava}, videos~\cite{team2023gemini} or 3D~\cite{zhu2024llava3d}, showing promise for robotic applications~\cite{driess2023palme,black2024pi0}.
In navigation specifically, early efforts have incorporated LLMs through prompt engineering and post-processed the models' text outputs~\cite{zhou2024navgpt,zhou2024navgpt2}. 
While achieving promising results, they fail to fully leverage rich visual information.
To address these limitations, subsequent research uses VLMs for navigation by processing the entire image history during navigation~\cite{zhang2024navid,zhang2024uninavid,chang2024mobilityvla}.
These remain computationally intensive due to the large number of visual tokens.
In this work, we integrate a compact, structured memory representation into VLMs to perform efficient exploration and navigation.

\section{Method}
\label{sec:method}
\vspace{-0.5em}

\subsection{Overview}
\label{subsec:task-definition}

\noindent{\textbf{Task definition.}}
The ObjectNav task~\cite{batra2020objectnav} requires an agent to navigate in an unknown indoor environment to find a specified object category such as a {\em ``chair"}.
At each step, the agent receives an RGB-D observation and its camera pose, assumed to be provided by an external method, \eg visual SLAM~\cite{thrun2008slambook}. It can take one of four actions: move forward by 25cm, turn left or right by 30°, or stop.
An episode is considered successful if the agent stops within 1 meter of an instance of the target object category, with the object visible from the stopping location, and within a 500-step limit.

\noindent{\textbf{Overall framework.}}
The proposed FOM-Nav model consists of three modules as illustrated in~\Cref{fig:overview}: online frontier-object mapping, high-level goal prediction, and low-level path planning.
The \emph{online mapping module} employs a segmentation model to process the RGB image at each step, projects the segmentation results into 3D space, and integrates them with the previously constructed map.
The process generates an updated frontier-object map, containing both segmented objects and newly identified frontiers which represent the boundary between explored and unexplored regions.
The \emph{goal prediction module} then selects either a frontier or an object as the navigation goal. 
Finally, the \emph{path planning module} generates an executable trajectory using the agent's four actions to reach the predicted high-level goal.

\vspace{-0.25em}
\subsection{Online Frontier-Object Mapping}
\label{subsec:map-plan}
\vspace{-0.25em}

We construct sparse 3D point clouds for obstacles and objects on-the-fly during navigation.
At each step $t$, the scene point cloud $P_s^t$ is used to build a 2D obstacle map and extract frontiers, while a separate point cloud $P_{o_i}^t$ for each object $o_i$ allows dynamically merging seen objects and appending newly discovered ones.

\vspace{0.25em}
\noindent{\textbf{Obstacle mapping.}}
Given the depth image and the camera pose, we back-project the depth image into a point cloud $\hat{P}^t_s$ expressed in the world coordinate frame, without considering semantic information.
We then add $\hat{P}^t_s$ to the previous scene point cloud $P^{t-1}_s$, and downsample it to a 5cm voxel resolution to obtain the current scene point cloud $P^{t}_s$.
From $P^{t}_s$, we generate a 2D obstacle and exploration map $M^t_s$ by projecting $P^{t}_s$ onto the horizontal plane, centering at the robot's current position, and inflating the obstacles by the robot's radius.
Using $M^t_s$, we follow prior work~\cite{yokoyama2024vlfm,yamauchi1997frontier} to track the explored regions and extract frontiers $\{F^t_j\}_{j=1}^{n_f}$. 
Additionally, we extract a visual feature $v^t_{f_j}$ for each newly detected frontier $j$ at step $t$, as detailed in~\Cref{subsec:llava-nav}.

\vspace{0.25em}
\noindent{\textbf{Object mapping.}}
For each RGB image at step $t$, we use an instance segmentation model Mask2Former~\cite{cheng2022mask2former} to automatically obtain segmentation masks with the corresponding probability distribution $\hat{c}_{o_i}^t$ over a set of predefined categories. 
For each segmented object $o_i$, we extract its visual feature $\hat{v}^t_{o_i}$ as detailed later in~\Cref{subsec:llava-nav}, and back-project its pixels using the depth image and camera pose to create a 3D point cloud.
We further filter isolated points in this point cloud using DBSCAN~\cite{ester1996dbscan} and store it at a low resolution of 5cm as $\hat{P}^{t}_{o_i}$.
At the first step, we convert all the segmented objects $\{\hat{P}^{1}_{o_i}\}$ into the initial object point clouds $\{P^1_{o_i}\}$.
In subsequent steps, for each newly segmented object, we first identify the closest existing object $P^{t-1}_{o_k}$ in $\{P^{t-1}_{o_i}\}$.
If the proportion of matched points within a small distance exceeds a defined threshold, we merge the new object $\hat{P}^{t}_{o_i}$ with $P^{t-1}_{o_k}$; otherwise, we add the object $\hat{P}^{t}_{o_i}$ as a new entity.
When merging objects, we combine their respective point clouds, compute a weighted average of their predicted category distributions and visual features, with weights proportional to the number of points in each point cloud.
If two objects are merged, we continue merging the resulting one with other objects in $\{P^{t-1}_{o_i}\}$ until no further merges are found. This is because merged objects may connect two original disconnected point clouds.
In this way, we obtain object point clouds $\{P^t_{o_i}\}_{i=1}^{n_o}$ from the entire observation history, where each object contains an accumulated visual feature $v^t_{o_i}$ and category distribution $c_{o_i}^t$.

\begin{figure*}[t]
    \centering
    \includegraphics[width=\linewidth]{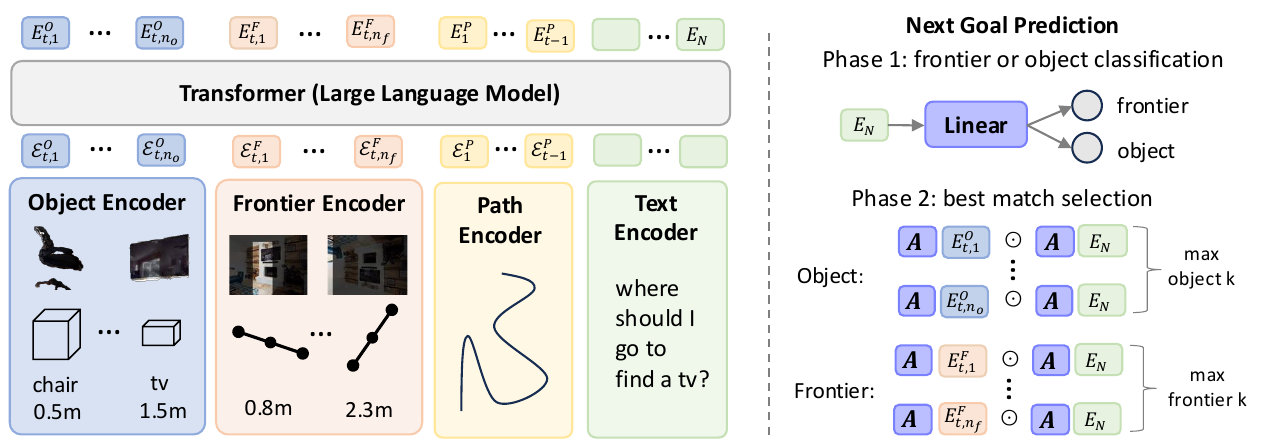}
    \vspace{-2em}
    \caption{
    Our high-level goal prediction module employs a transformer-based LLM to processes a rich semantic representation of the scene, including objects, frontiers, path history and texts. Objects and frontiers consist of visual, geometric, distance, and textual features.
    The prediction head operates in two phases: 1) the transformer's last output token $E_N$ is classified as either a frontier or object destination type; 2) $E_N$ is matched against output tokens for objects or frontiers using a learnable matrix $A$ to identify the best navigation goal.
    }
    \label{fig:large_model}
    \vspace{-2em}
\end{figure*}

\vspace{-0.25em}
\subsection{High-level Goal Prediction}
\label{subsec:llava-nav}
\vspace{-0.25em}

Our high-level goal prediction model adopts a similar architecture to recent VLMs including LLaVA~\cite{liu2023llava}.
As illustrated in \Cref{fig:large_model}, the model processes the frontier-object map to determine the next navigation goal.

\vspace{0.25em}
\noindent{\textbf{Input Encoding.}}
We encode frontiers, objects and the robot's history path separately.
All the spatial coordinates below are expressed in the agent's egocentric frame.

\emph{\textbf{Frontier embedding:}} 
It combines three complementary features: geometric, visual and distance features.

The geometric feature encodes the 2D spatial configuration of the frontier in the obstacle map $M^t_s$.
Let $F^t_{j}$ denote a frontier at step $t$, represented as a boundary line from $(x^{e_1}_j, y^{e_1}_j)$ to $(x^{e_2}_j, y^{e_2}_j)$, with center point $(x^c_j, y^c_j)$.
We use two multi-layer perceptrons (MLP) $f^F_{ends}$ and $f^F_{center}$ to encode the endpoints and the center point respectively.

The visual feature captures the surrounding environment of the frontier.
Let $I^{t'}$ represent the RGB image at which the frontier $F^t_{j}$ is first detected. 
We apply DINO-v2~\cite{oquab2024dinov2,darcet2024registers} to extract patch-level image features from $I^{t'}$.
Instead of using a global image feature, we extract localized visual features around the frontier.
Given all 2D points along the frontier line, we extend them vertically to 3D points with heights ranging from ground level to 50cm above the robot. These 3D points are projected onto the camera frame of $I^{t'}$, generating a rectangular mask for the frontier region in the image.
The frontier mask is resized to match the patch-level feature size, and we compute the mean feature within the masked region to obtain the frontier's visual feature $v^t_{f_j}$, which is then processed by another MLP, $f_{vis}^{F}$.

The distance feature quantifies the cost of reaching a frontier. We compute the geodesic distance $d_j$ from the robot's current position to the frontier center on the obstacle map $M^t_s$ using the fast marching method (FMM)~\cite{sethian1996fastmarching}. This distance is encoded using an MLP, $f_{dist}^{F}$.

The final embedding of the frontier $F^t_j$ is computed as follows, where $T_F$ a learnable vector representing the type of frontier tokens:
\begin{align}
\begin{split}
    \mathcal{E}^{F}_j &= f^F_{center}(x^c_j, y^c_j) + f^F_{ends}(x^{e_1}_j,y^{e_1}_j,x^{e_2}_j,y^{e_2}_j) \\
    & + f^F_{vis}(v_{f_j}) + f^F_{dist}(d_j) + T_F.
\end{split}
\end{align}

\emph{\textbf{Object embedding:}}
Similar to frontier embedding, object embedding integrates geometric, distance, visual and additional text category features.

The geometric feature encodes the object's spatial information, including the coordinates of its point cloud center $(x^t_{o_i},y^t_{o_i},z^t_{o_i})$ and bounding box corner coordinates $B^t_{o_i}$.
The distance feature encodes the geodesic distance $d_{o_i}$ from the robot's current position to the object.

The visual feature is accumulated over all frames in which the object is detected according to the merged object point cloud $P^t_{o_i}$.
At each frame, we extract patch-level image features using Dino-v2. The object's segmentation mask is used to average patch-level features within the masked region as $\hat{v}^{t}_{o_i}$. 
The final visual feature $v^{t}_{o_i}$ is a weighted average of $\{\hat{v}^{t'}_{o_i}\}_{t'=1}^{t}$, providing robustness to partial occlusions and varying viewpoints.

The text category feature is derived from the object’s accumulated category distribution $c^t_{o_i}$.
We select the category name with the highest probability in $c^t_{o_i}$ and encode it using the CLIP~\cite{radford2021clip} text encoder $f_{cat}^{O}$.

The final embedding of the object $o_i$ is as follows:
\begin{equation}
\begin{split}
    \mathcal{E}^{O}_t &= f^O_{center}(x^t_{o_i},y^t_{o_i},z^t_{o_i}) + f^O_{corners}(B^t_{o_i}) \\
    &+ f^O_{dist}(d_{o_i}) + f^O_{vis}(v^{t}_{o_i}) + f^O_{cat}(c^t_{o_i}) + T_O,
\end{split}
\end{equation}
where $T_O$ is a learnable vector for denoting the type of the object token.

\emph{\textbf{Path embedding:}}
The history path can provide contextual information to maintain decision consistency for navigation. 
We encode 2D locations $(x_{t'},y_{t'})_{t'=1}^{t}$ expressed in the current robot frame along the robot's history path via an MLP $f^P_{\text{loc}}$ as follows, with $T_P$ being a learnable vector to represent the type of path tokens:
\begin{align}
    \mathcal{E}^{P}_{t'} &= f^P_{\text{loc}}(x_{t'},y_{t'}) + T_P.
\end{align}

\vspace{0.25em}
\noindent{\textbf{LLM.}}
We construct a prompt for the LLM with the following template:
``Given seen objects \texttt{<objects>}, frontiers of explored areas \texttt{<frontiers>} and past locations \texttt{<path history>}, where should I go to find a \texttt{<target category>}?"
The \texttt{<target category>} is replaced with the specific category name to search for. The \texttt{<objects>}, \texttt{<frontiers>} and \texttt{<path history>} are substituted with their corresponding token embeddings.
Early experiments with other prompts showed similar performance.

Given $n_f$ frontier tokens and $n_o$ object tokens, the LLM generates corresponding output embeddings for frontiers and objects, denoted as $E_N \in \mathbb{R}^{n_f \times D}$ and $E_O \in \mathbb{R}^{n_o \times D}$ respectively, where $D$ is the feature dimensionality.
Additionally, the model produces an embedding for the next token prediction, represented as $E_N \in \mathbb{R}^{1 \times D}$.

\vspace{0.25em}
\noindent{\textbf{Prediction Head.}}
We predict the next navigation goal in two phases. 
First, we apply a binary classification to the next token embedding $E_N$ using a linear layer to determine whether the next goal should be a frontier or an object. 
Based on this decision, we compute the similarity between $E_N$ and the frontier or object embeddings as:
\begin{align}  
    S =  E_N^T (A^TA) E,
\end{align}
where $A\in\mathbb{R}^{D\times D'}$ is a learnable matrix and $E=E_F$ if $E_N$ is classified as a frontier and $E_O$ otherwise.
The next goal is selected as the frontier or object with the highest similarity.

\vspace{0.25em}
\noindent{\textbf{Training.}}
We train the navigation model on a dataset described in~\Cref{sec:dataset}. 
The training objective consists of two components: a binary cross-entropy (BCE) loss $\mathcal{L}_{BCE}$ for frontier-object classification and a cross-entropy (CE) loss $\mathcal{L}_{CE}$ for selecting the specific frontier or object.
\begin{equation}
    \mathcal{L} = \mathcal{L}_{BCE} + \lambda \mathcal{L}_{CE}.
\end{equation}

\subsection{Low-level Path Planning}
\label{subsec:path-plan}

Given the predicted goal from the high-level model, we first determine a precise target location or area. 
We then use a path planner to navigate towards this target.

\vspace{0.25em}
\noindent{\textbf{Determining the target location.}}
The robot follows different strategies based on the goal type. 
If the next goal is a frontier, we consider the center point of the frontier as the target. 
If the goal is an object, we either compute a valid stopping area next to it if the object is in the explored area, or we move to the closest frontier to the target.

\vspace{0.25em}
\noindent{\textbf{Path planner.}}
Once the target location/area is determined, we compute a distance map from that location to the explored area using the Fast Marching algorithm (FMM)~\cite{sethian1996fastmarching}, which serves as a heuristic cost for A${}^*$~\cite{hart1968astar}.
Unlike prior work~\cite{chaplot2020semexp,luo2022stubborn} using FMM alone, incorporating A${}^*$ on top of FMM ensures consistent long-term trajectory planning.
Following~\cite{chaplot2020semexp,luo2022stubborn}, we maintain a collision map that stores locations where the agent collides with the environment, defined as no position change after a {\em move forward} action, or gets stuck in holes of the scanned environment meshes.
This collision map is combined with the obstacle map to extract frontiers and plan a path.
As in prior work~\cite{yokoyama2024vlfm}, we start each episode with a 360° rotation to observe the surroundings before taking any action.

Compared to point-goal navigation policies~\cite{wijmans2020ddppo} used in prior works~\cite{yokoyama2024vlfm}, our path planner is not restricted to discrete action spaces, making it directly applicable to navigation with continuous actions.
In that case, A${}^*$ becomes unnecessary and collision maps are less important, only restricted to scene parts invisible to a fixed camera.

\section{ObjectNav Dataset Construction}
\label{sec:dataset}

Existing ObjectNav datasets either use shortest paths~\cite{ramakrishnan2021hm3d}, which limit exploration, or rely on costly human demonstrations~\cite{ramrakhya2022habitatweb}. To overcome this, we automatically generate navigation trajectories using our online frontier–object maps, enabling scalable training for FOM-Nav.

\noindent{\textbf{Dataset construction.}}
Our dataset is built upon scanned real-world environments with semantic annotations such as HM3D~\cite{ramakrishnan2021hm3d} and MP3D~\cite{chang2017mp3d}.

We first pre-process each scene to obtain ground-truth stopping locations for each object. These are viewpoints where the target object is visible and within 1 meter, consistent with the evaluation metric.
We then collect navigation trajectories using the Habitat simulator~\cite{savva2019habitat}, randomly selecting initial robot locations and target objects on the same floor, as crossing stairs with wheeled robots is impractical.

At the start of each navigation episode, the robot performs a 360° rotation to observe its surroundings.
It constructs a frontier-object map on-the-fly as described in~\ref{subsec:map-plan}.
At each step, in the case a target object is determined to be visible, it is selected as the ground-truth next sub-goal.
Otherwise, for each frontier, we compute its distance to the robot and to the closest ground-truth stopping locations. 
The best frontier is chosen as the one minimizing the average distance to the agent and the target.

We collect datasets using either ground-truth segmentations or automatic ones from an instance segmentation model for online mapping.
In the latter case, we still rely on ground-truth segmentations to identify the target object in the automatically constructed object maps, which is the object with highest point cloud overlap with the ground-truth object.
We store the built object maps, frontiers, path history and target labels at each step.

\vspace{0.25em}
\noindent{\textbf{Dataset statistics.}}
We collect navigation trajectories with target categories in existing benchmarks on training environments.
For the HM3D v1/v2 benchmark~\cite{ramakrishnan2021hm3d}, we gather a total of 490k and 270k steps with either ground-truth or automatic segmentations for mapping from 145 scenes, corresponding to 12.6k and 7k episodes, respectively.
For the MP3D benchmark~\cite{chang2017mp3d}, we use 50 scenes and collect 215k and 80k steps with ground-truth or automatic segmentation in a total of 5k and 2k episodes, respectively.

\section{Experiments}
\label{sec:experiments}

\subsection{Experimental Setup}

\noindent{\textbf{ObjectNav Benchmarks.}}
We evaluate our model on three ObjectNav benchmarks.
The MP3D benchmark~\cite{batra2020objectnav} evaluates 21 object categories, and has 56 training and 11 validation scenes.
The HM3D benchmark~\cite{ramakrishnan2021hm3d} focuses on 6 object categories, including 145 scenes for training and 36 for validation.
It contains two validation versions: v1 from Habitat 2022 challenge~\cite{habitatchallenge2022} and v2 from Habitat 2023 challenge~\cite{habitatchallenge2023}.
The main difference is that v2 has better ground-truth annotations and does not require navigation across floors.
The HM3D-OVON benchmark~\cite{yokoyama2024hm3dovon} operates on the same HM3D environments but includes hundreds of target categories, split between those seen, unseen or synonyms to objects in training scenes.

\vspace{0.25em}
\noindent{\textbf{Evaluation metrics.}}
We use two primary metrics following~\cite{batra2020objectnav} for online navigation evaluation: success rate (SR) and success rate weighted by path length (SPL).
The SR metric measures the ratio of successful episodes where the robot stops within 1 meter to the target object with the object visible, and within 500 steps.
The SPL metric considers the robot's path length $d_T$ and the geodesic distance $d_G$ from the start to the closest target object, defined as: $SPL = \text{Success} \times \frac{d_G}{max(d_G, d_T)}$.
This metric emphasizes exploration efficiency besides accuracy.

\vspace{0.25em}
\noindent{\textbf{Implementation details.}}
We train FOM-Nav using the constructed dataset described in~\Cref{sec:dataset} for each benchmark.
For online mapping, we train Mask2Former~\cite{cheng2022mask2former} on images from training scenes of the respective benchmarks as the instance segmentation model.
For the high-level goal prediction model, the encoding MLPs all consist of 2 hidden layers with 2048 neurons per layer.
The LLM is initialized from LLaVA-v1.6 7B model~\cite{liu2023improvedllava} and remains frozen.
The decoding projection in the prediction head has $D'=2048$.
The goal prediction model is trained for 2 epochs on 4 H100 GPUs, which takes around 2 hours for MP3D and 4 hours for HM3D, using a learning rate of $2 \times 10^{-4}$ and batch size of 32 per GPU.
We use $\lambda=0.5$ in the loss.

\vspace{-0.2em}
\subsection{Ablation studies}
\vspace{-0.3em}

\begin{table}
\centering
\caption{Ablation of input encoding for high-level goal prediction on the HM3D v2 validation split.}
\vspace{-0.5em}
\label{tab:ablation_input_encoding}
\begin{tabular}{lcccccc} \toprule
\multirow{2}{*}{} & \multirow{2}{*}{\begin{tabular}[c]{@{}c@{}}Frontier\\ image\end{tabular}} & \multicolumn{2}{c}{Object feature} & \multirow{2}{*}{\begin{tabular}[c]{@{}c@{}}Path\\ tokens\end{tabular}} & \multirow{2}{*}{SR} & \multirow{2}{*}{SPL} \\
 &  & Image & Text &  & & \\ \midrule
(a) & \xmark & \xmark & \cmark & \cmark & 63.7 & 38.2 \\
(b) & Global & \xmark & \cmark & \cmark & 63.8 & 38.7 \\
(c) & Global & \cmark & \cmark & \cmark & 70.5 & 42.2 \\
(d) & Local & \xmark & \cmark & \cmark & 70.3 & 45.5 \\
(e) & Local & \cmark & \xmark & \cmark & 72.6 & 46.5 \\
(f) & Local & \cmark & \cmark & \xmark & 74.5 & 47.4 \\
(g) & Local & \cmark & \cmark & \cmark & \textbf{75.8} & \textbf{47.9} \\ 
\bottomrule
\end{tabular}
\vspace{-0.5em}
\end{table}

\noindent{\textbf{High-level goal prediction.}}
\Cref{tab:ablation_input_encoding} presents ablations on input encodings of our high-level goal prediction module. 
Variant (g) is our final model using localized image features for frontiers, both image and text features for objects and history path tokens as inputs.
First, we evaluate the impact of image features for both frontiers and objects.
The variant (a) omits all image features for frontiers or objects, yielded the worst performance.
Incorporating global image features into frontier embeddings, as shown in variant (b), slightly improves the performance.
Further gains are observed when object image features are integrated as demonstrated in variant (c). 
Comparing variants (b) and (d) or (c) and (g), replacing global frontier image features with more precisely localized image features substantially boosts performance, \eg, with 5.3 and 5.7 improvement points on SR and SPL respectively from (c) to (g).
Next, we show that text features are complementary to image features in object embeddings, as seen from variants (e) and (g).
Finally, including path tokens leads to a slight performance increase comparing variants (f) and (g).

\begin{table}
\centering
\caption{
Ablation of different path planners with our method on the HM3D v2 validation split.
}
\vspace{-0.5em}
\label{tab:path_planning}
\begin{tabular}{lcc} \toprule
& SR & SPL \\ \midrule
 Habitat GT greedy geodesic & \textbf{77.1} & 47.6 \\
PointNav~\cite{anderson2018evaluation} & 75.1 & 45.0 \\
 A${}^{*}$~\cite{hart1968astar} + FMM~\cite{sethian1996fastmarching} & 75.8 & \textbf{47.9} \\
\bottomrule
\end{tabular}
\vspace{-1.5em}
\end{table}

\begin{table}[t]
\centering
\caption{Ablations of the training data size and the use of ground-truth (GT) or automatic (Auto) frontier-object maps during data collection on the HM3D v2 validation split.
}
\vspace{-0.5em}
\label{tab:ablation_training_data}
\begin{tabular}{lcccccc} \toprule
 & \multicolumn{2}{c}{Training size} & \multicolumn{2}{c}{FO Maps} & \multirow{2}{*}{SR} & \multirow{2}{*}{SPL} \\
 & Envs & Steps & GT & Auto &  &  \\ \midrule
(a) & 70 & 200k & \cmark & \cmark & 70.7 & 46.2 \\
(b) & 145 & 200k & \cmark & \cmark & 70.6 & 45.2 \\
(c) & 145 & 490k & \cmark & \xmark & 61.7 & 37.5 \\
(d) & 145 & 270k & \xmark & \cmark & 73.6 & 46.5 \\
(e) & 145 & 760k & \cmark & \cmark & \textbf{75.8} & \textbf{47.9} \\ \bottomrule
\end{tabular}
\vspace{-0.5em}
\end{table}

\vspace{0.25em}
\noindent{\textbf{Low-level path planning.}}
We compare different path planners in our low-level path planning module in~\Cref{tab:path_planning}.
The ground-truth path planner in Habitat, based on the environment GT map, gives the best SR. 
Our proposed path planner, while decreasing the SR compared to ground truth, achieves comparable SPL.
This performance is notably better than the PointNav policy trained in these environments, demonstrating that our planner navigates more efficiently.
In addition, our path planner is more flexible and directly applicable to the real world given obstacle maps, while PointNav policies require re-training.

\vspace{0.25em}
\noindent{\textbf{Training data.}}
\Cref{tab:ablation_training_data} evaluates the impact of training data size on our model's performance.
Our final model in (e) utilizes 760k steps across 145 environments, including 490k data constructed with ground-truth segmentations and 270k data with automatic segmentation.
Comparing variants (c) and (d) with (e), we can see that the data with automatic segmentation plays a more important role than the data with ground-truth one even if the size is smaller, as it helps reduce the gap between training and inference.
But both data are complementary and the ground-truth one is more computationally efficient to obtain as it requires no segmentation models.
In variant (b), we reduce the dataset size to 25\% of the total episodes still across 145 environments, and observe a performance drop compared to (e).
Finally, variant (a), which uses the same amount of 200k steps but only in 70 environments, has a similar decrease in performance, indicating similar importance of the number of trajectories and the diversity of environments in training.

\noindent{\textbf{Image segmentation.}}
When using ground-truth segmentation masks during our frontier-object mapping, our method reaches 86.6\% SR and 56.1 points for the SPL.
We observe a drop of 11.0\% for SR and 8.2 points for SPL when moving to our automatic instance segmentation.
Failures of the automatic segmentation are mainly due to objects being segmented with wrong labels, or objects never being segmented.
This shows that a large improvement is expected by integrating stronger instance segmentation models, either image- or video-based ones.

\subsection{Comparison to the state of the art}

\begin{table}
\centering
\small
\caption{
Results on the validation splits of MP3D$_{sub}$, HM3D v1$_{sub}$ and HM3D v2, where we exclude episodes requiring cross-floor navigation for MP3D and HM3D v1.
VLFM${}^*$ is our reproduced results and VLFM${}^{\dagger}$ is VLFM with our  segmentation model.
}
\label{tab:sota_cmpr_single_floor}
\vspace{-0.75em}
\begin{tabular}{lcccccc} \toprule
\multirow{2}{*}{} & \multicolumn{2}{c}{MP3D$_{sub}$} & \multicolumn{2}{c}{HM3D v1$_{sub}$} & \multicolumn{2}{c}{HM3D v2} \\ 
 & SR & SPL & SR & SPL & SR & SPL \\ \midrule
RIM*~\cite{chen2023rim}
& \textbf{51.8} & 15.8 & 63.0 & 27.6 & 58.4 & 22.2 \\
PIRLNav*~\cite{ramrakhya2023pirlnav}
& --- & --- & 71.6 & 34.7 & 66.0 & 27.0 \\
VLFM*~\cite{yokoyama2024vlfm}
& 40.6 & 19.6 & 64.7 & 37.6 & 64.0 & 33.0 \\
VLFM${}^{\dagger}$~\cite{yokoyama2024vlfm}
& 42.8 & 19.8 & 65.9 & 39.6 & 60.3 & 33.6 \\
FOM-Nav
& 44.6 & \textbf{23.9} & \textbf{73.0} & \textbf{52.1} & \textbf{75.8} & \textbf{47.9} \\ \bottomrule
\end{tabular}
\vspace{-1em}
\end{table}

\begin{table}[t]
\centering
\small
\caption{Results on the validation splits of MP3D and HM3D v1 benchmarks including episodes that require crossing floors. VLFM${}^*$ is our reproduced results and VLFM${}^{\dagger}$ is VLFM with our  segmentation model.}
\label{tab:sota_cmpr_full}
\vspace{-0.5em}
\begin{tabular}{llcccc} \toprule
\multirow{2}{*}{Memory} & \multirow{2}{*}{Method} & \multicolumn{2}{c}{MP3D} & \multicolumn{2}{c}{HM3D v1} \\
& & SR & SPL & SR & SPL \\ \midrule
\multirow{5}{*}{Implicit} & Habitat-Web~\cite{ramrakhya2022habitatweb} & 35.4 & 10.2 & 57.6 & 23.8 \\
& RIM~\cite{chen2023rim} & \textbf{50.3} & \textbf{17.0} & 57.8 & 27.2 \\
& OVRL-v2~\cite{yadav2023ovrlv2} & --- & --- & 64.7 & 28.1 \\
& PIRLNav~\cite{ramrakhya2023pirlnav} & --- & --- & 70.4 & 34.1 \\
& XGX~\cite{wasserman2024xgx} & --- & --- & \textbf{72.9} & \textbf{35.7} \\
\midrule
\multirow{5}{*}{Explicit}
& SGM~\cite{zhang2024imaginebeforego}
& 37.7 & 14.7 & \textbf{60.2} & 30.8 \\
& VLFM~\cite{yokoyama2024vlfm} & \textbf{36.4} & 17.5 & 52.5 & 30.4 \\
& VLFM${}^*$~\cite{yokoyama2024vlfm}
& 34.4 & 16.7 & 52.0 & 30.2 \\
& VLFM${}^{\dagger}$
& 36.3 & 17.1 & 53.5 & 32.0 \\
& FOM-Nav (Ours)
& 35.0 & \textbf{18.7} & 57.3 & \textbf{40.8} \\ \bottomrule
\end{tabular}
\vspace{-1.5em}
\end{table}

\noindent{\textbf{MP3D and HM3D benchmarks.}}
We compare our method with the state-of-the-art methods VLFM, RIM~\cite{chen2023rim} and PIRLNav~\cite{ramrakhya2023pirlnav} in~\Cref{tab:sota_cmpr_single_floor} on MP3D and HM3D benchmarks excluding episodes that require crossing floors.
Our reproduced results are denoted with a *.
For fair comparison, we propose an improved baseline, VLFM${}^{\dagger}$, which is VLFM but using our instance segmentation model.
On MP3D$_{sub}$, our method obtains higher SPL than all the baselines, and is only outperformed by RIM in the SR. 
Note that GT annotations in these environments are often incomplete, leading to numerous episodes being considered as failures whereas a target object was actually reached.
On HM3D v1/v2 benchmarks with higher-quality annotations, the performance improvement is larger. For example, FOM-Nav achieves an increase of up to 11.8 and 14.3 points in SR and SPL respectively compared to the best VLFM on HM3D v2.

We include for completeness in~\Cref{tab:sota_cmpr_full} results obtained on the full MP3D and HM3D v1 benchmarks including those cross-floor navigation episodes.
Implicit memory-based approaches are able to navigate through stairs given cross-floor training data, while explicit ones currently consider them as obstacles since wheeled robots cannot climb stairs. 
In this setting, implicit methods do, as expected, better.
Identifying stairs and considering them as traversible areas connecting two floors would address this limitation. 
Compared to the state of the art, our method has a SR comparable to the best explicit approaches and to implicit ones such as Habitat-Web~\cite{ramrakhya2022habitatweb} on both benchmarks.
Notably, it obtains the highest SPL metric among all the methods, showcasing again the efficiency of our exploration.

\noindent{\textbf{OVON benchmark.}}
To assess open-vocabulary ObjectNav, we further evaluate on the OVON benchmark in~\Cref{tab:ovon}. For efficiency, we use Llama 3.2 1B~\cite{dubey2024llama3} as the base LLM. 
Because FOM-Nav introduces its own input encoders and output heads, it cannot directly inherit the generalization capabilities of pretrained VLMs without training these components on Internet-scale data. We therefore evaluate two variants of our model.
The first variant is trained on all annotated categories in the HM3D training houses (row HM-all), including synonyms and unseen categories within those houses. This model achieves a large SPL improvement over prior methods, demonstrating its efficient open-vocabulary navigation ability. Its SR is comparable to VLFM and DAgRL+OD but lower than Uni-NaVid and MTU3D, suggesting that Internet-scale pretraining is important for object recognition. 
However, this setup is somewhat favorable to us since unseen categories appear in the training environments, whereas prior methods rely solely on Internet data for generalization to unseen categories.
To create a fairer comparison, we keep episodes targeting synonyms and unseen categories but remove their ground-truth labels, automatically annotating targets using a VLM (row HM-all auto-label). Although performance decreases relative to HM-all, this variant still surpasses or achieves comparable performance to previous methods in SPL, using far fewer resources than Uni-Navid and MTU3D.
Future work may improve generalization further by training FOM-Nav on larger-scale vision–language data and more environments.

\begin{table}
\centering
\small
\tabcolsep=0.125cm
\caption{
Results on OVON validation splits. 
VLFM~\cite{yokoyama2024vlfm} and DAgRL+OD~\cite{yokoyama2024hm3dovon} use OWLv2~\cite{minderer2023owlv2} open-vocabulary detector. 
Uni-Navid~\cite{zhang2024uninavid} and MTU3D~\cite{zhu2025mtu3d} train on large-scale vision-language data and navigation data. `HM-all' uses all HM3D categories and 
`auto-label' replaces GT labels for synonyms and unseen categories with automatic annotations.
}
\label{tab:ovon}
\vspace{-0.75em}
\scalebox{0.9}{
\begin{tabular}{cccccccc} \hline
\multirow{2}{*}{Method} &  \multirow{2}{*}{\begin{tabular}[c]{@{}c@{}}Training \\ categories\end{tabular}} & \multicolumn{2}{c}{Val seen} & \multicolumn{2}{c}{\begin{tabular}[c]{@{}c@{}}Val synonyms\end{tabular}} & \multicolumn{2}{c}{Val unseen} \\ 
 &  & SR & SPL & SR & SPL & SR & SPL \\ \hline
VLFM & open-vocab & 35.2 & 18.6 & 32.4 & 17.3 & 35.2 & 19.6 \\
DAgRL+OD & open-vocab & 38.5 & 21.1 & 39.0 & 21.4 & 37.1 & 19.9 \\ 
Uni-Navid & open-vocab & 41.3 & 21.1 & 43.9 & 21.8 & 39.5 & 19.8 \\
MTU3D & open-vocab & \textbf{55.0} & 23.6 & \textbf{45.0} & 14.7 & \textbf{40.8} & 12.1 \\
\hline
\multirow{2}{*}{\begin{tabular}[c]{@{}c@{}}FOM-Nav\\ (Ours)\end{tabular}}
& HM-all & 42.5 & \textbf{27.8} & 36.3 & \textbf{23.4} & 38.1 & \textbf{25.7} \\
& \begin{tabular}[c]{@{}c@{}}HM-all auto\end{tabular}
& 37.6 & 24.6 & 29.4 & 18.5 & 30.9 & 20.6 \\ \hline
\end{tabular}
}
\vspace{-0.75em}
\end{table}

\begin{figure}
    \centering
    \includegraphics[width=\linewidth]{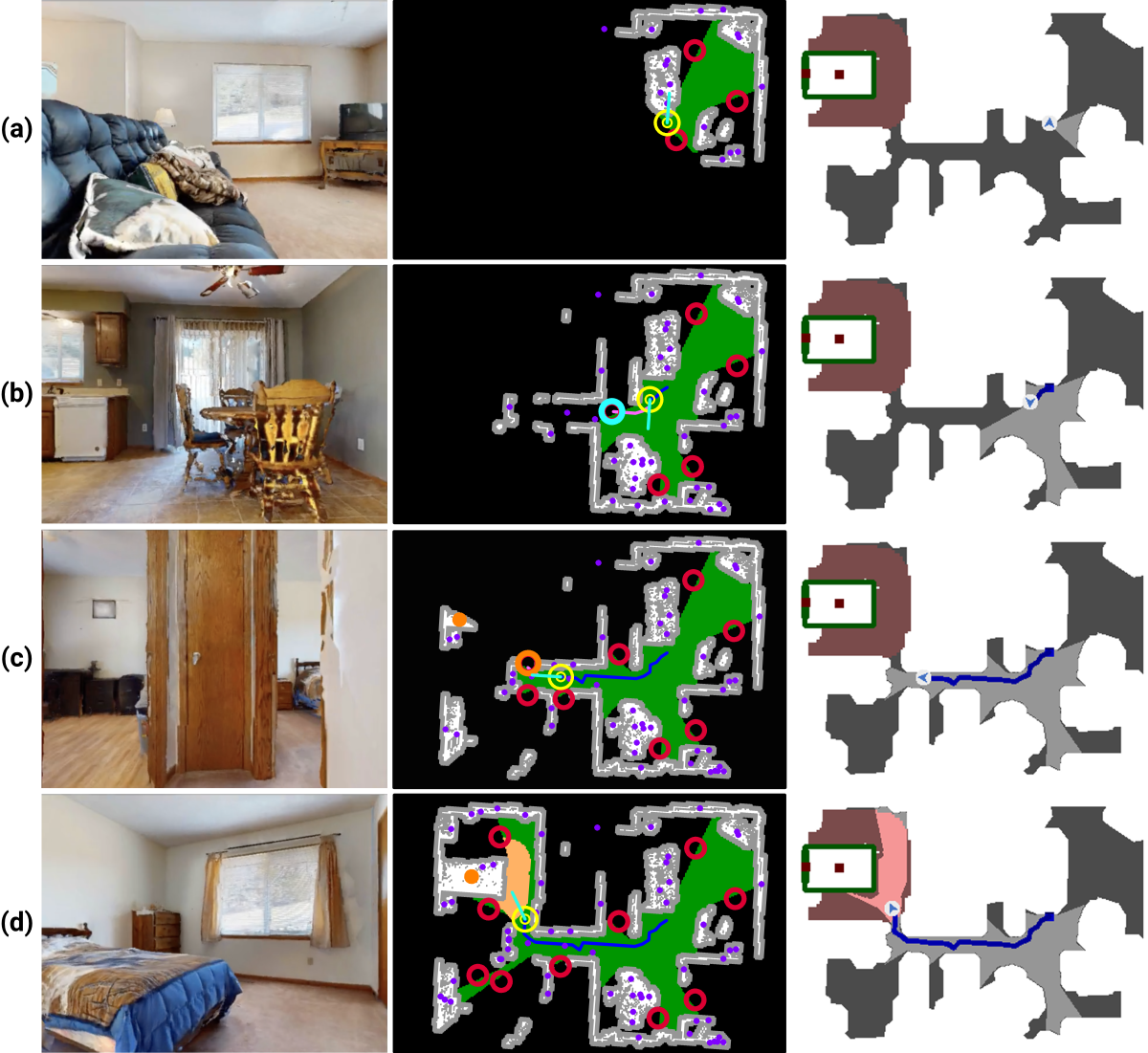}
    \vspace{-1.5em}
    \caption{
    Visualization of a navigation episode to a {\em bed} in an HM3D environment~\cite{ramakrishnan2021hm3d}.
    Left: RGB view from the agent.
    Center: Online frontier–object maps: grey = inflated obstacles, green = explored area, purple dots = objects, red circles = frontiers, and filled orange/cyan circles = selected object/frontier goals. If the chosen object lies in unexplored space, the agent first moves toward the corresponding frontier. Planned and executed paths are shown in pink and navy.
    Right: Agent trajectory (navy), ground-truth object (white box), and stopping area (pink) on the GT map.
    }
    \label{fig:traj-bed}
    \vspace{-1.65em}
\end{figure}

\noindent{\textbf{Qualitative results.}}
We show in~\Cref{fig:traj-bed} an example of an episode with the target category {\em bed}.
After an initial self-rotation (a), the agent decides to go to the corridor instead of searching in the kitchen (b).
Once the bed becomes visible, FOM-Nav selects it as the next target instead of exploring another frontier and room (c).
It then computes a valid stopping area and stops inside (d).

\subsection{Real-world deployment}

We deploy our model on a wheeled TIAGo++ robot in our laboratory.
We mount an Orbbec Femto Mega RGB-D camera on the robot's head.
All the models are run on an external desktop equipped with an RTX 4060 GPU.
The localization is done with the RGB-D version of ORB-SLAM3~\cite{campos2021orbslam3} and the image segmentation uses an off-the-shelf Mask2Former trained on the ADE20K dataset.
Our supplementary video demonstrates the successful transfer of our method to the real world.

\vspace{-0.2em}
\section{Conclusion}
\label{sec:conclusion}
\vspace{-0.2em}

This paper introduces FOM-Nav for object goal navigation, leveraging frontier-object maps for rich semantic scene representation, a vision-language model for goal prediction, and low-level path planner for action execution.
Trained on automatically constructed navigation datasets, FOM-Nav achieves state-of-the-art performance on two ObjectNav benchmarks, particularly in exploration efficiency.
Future work includes extending the approach to multi-floor and more complex semantic navigation tasks.

\section*{Acknowledgment}
This work was granted access to the HPC resources of IDRIS under the allocation 2021-AD011012725 made by GENCI.
It was supported in part by the French government under management of Agence Nationale de la Recherche as part of the “France 2030” program, PR[AI]RIE-PSAI projet, reference ANR-23-IACL-0008.
JP was supported in part by the Louis Vuitton/ENS chair in artificial intelligence and a Global Distinguished Professorship at the Courant Institute of Mathematical Sciences and the Center for Data Science at New York University.

\vspace{-0.25em}
\bibliographystyle{IEEEtran}
\bibliography{IEEEabrv,main}

\clearpage
\renewcommand{\thefigure}{S\arabic{figure}}
\setcounter{figure}{0}
\renewcommand{\thetable}{S\arabic{table}}
\setcounter{table}{0}
\renewcommand{\thesection}{S\arabic{section}}
\setcounter{section}{0}
\setcounter{page}{1}
% \maketitlesupplementary

\section{Additional implementation details}

\subsection{Instance segmentation model}
\noindent We fine-tune Mask2Former~\cite{cheng2022mask2former} pretrained on Ade20k dataset as our instance segmentation model.
To this end, we automatically collect 2000 RGB images taken at random view points in each training house of the MP3D~\cite{chang2017mp3d} and HM3D~\cite{ramakrishnan2021hm3d} benchmarks.
Ground-truth segmentation masks are rendered for each image using the Habitat simulator~\cite{savva2019habitat} and the annotated meshes from MP3D and HM3D~\cite{yadav2023hm3dsemantics}.
For MP3D, annotations contain around 35 categories.
For HM3D, we remove labels corresponding to walls, floors or ceilings and use the 1.4k remaining categories for segmentation.
The large diversity of labels in HM3D may provide more fine-grained scene understanding to learn richer environment priors and enable efficient navigation than the limited set of labels from MP3D.
Tiny objects occupying less than 100 pixels are excluded from training.
We then train one segmentation model per benchmark.
The model is fine-tuned for one epoch on the collected dataset, using a learning rate of $5 \times 10^{-5}$ with Adam optimizer and a batch size of 16. 
During online evaluations, we keep masks with a class probability larger than a threshold of 0.65 for HM3D and 0.8 for MP3D.
We filter masks that are too close to the borders of an image, \ie, at less than 30 pixels from the top or bottom or 20 pixels from the left or right sides, and discard masks covering less than 100 pixels.
We use the same models and parameters for the VLFM${}^{\dagger}$ baseline.

\subsection{Frontier and object mapping}

\begin{figure*}
    \centering
    \begin{subfigure}[b]{0.485\linewidth}
         \centering
         \includegraphics[width=\linewidth]{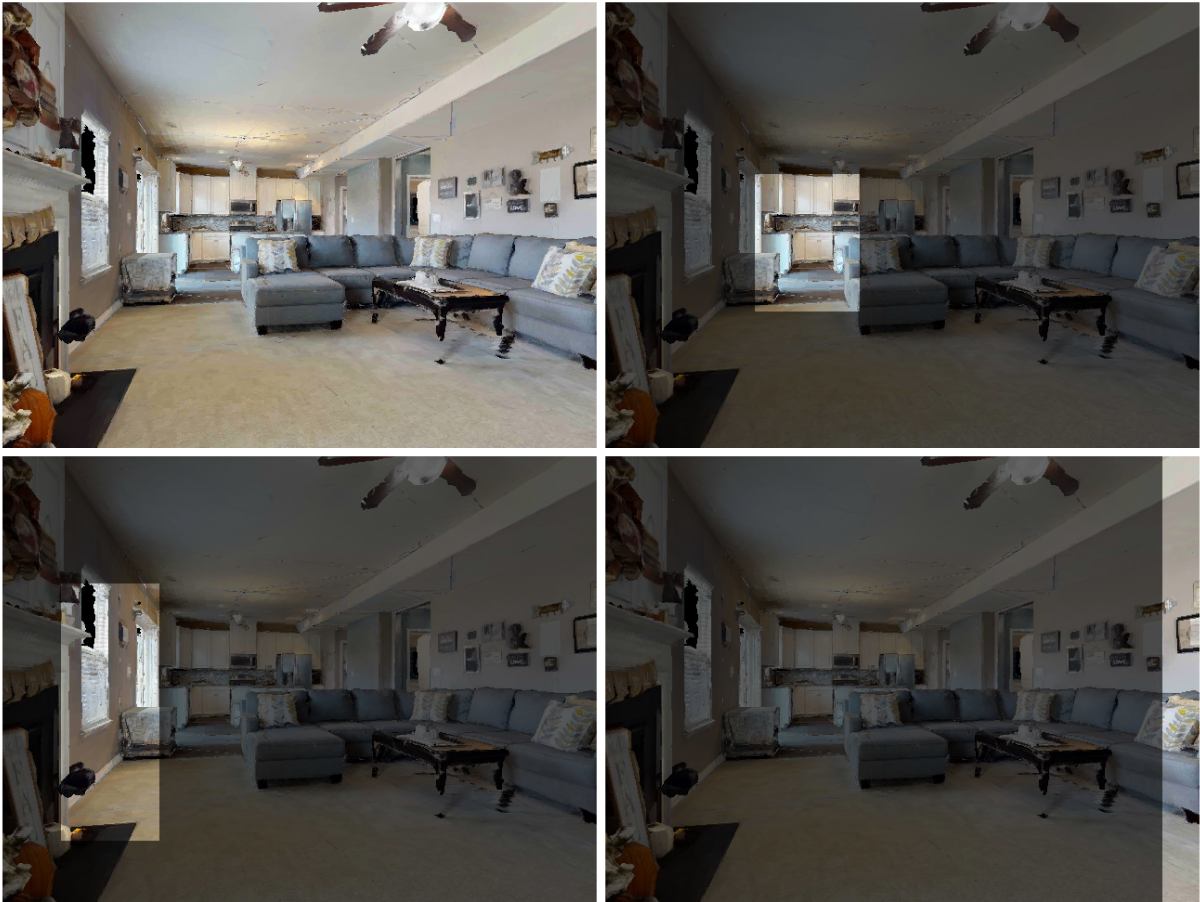}
    \end{subfigure}
    \hfill
    \begin{subfigure}[b]{0.485\linewidth}
         \centering
         \includegraphics[width=\linewidth]{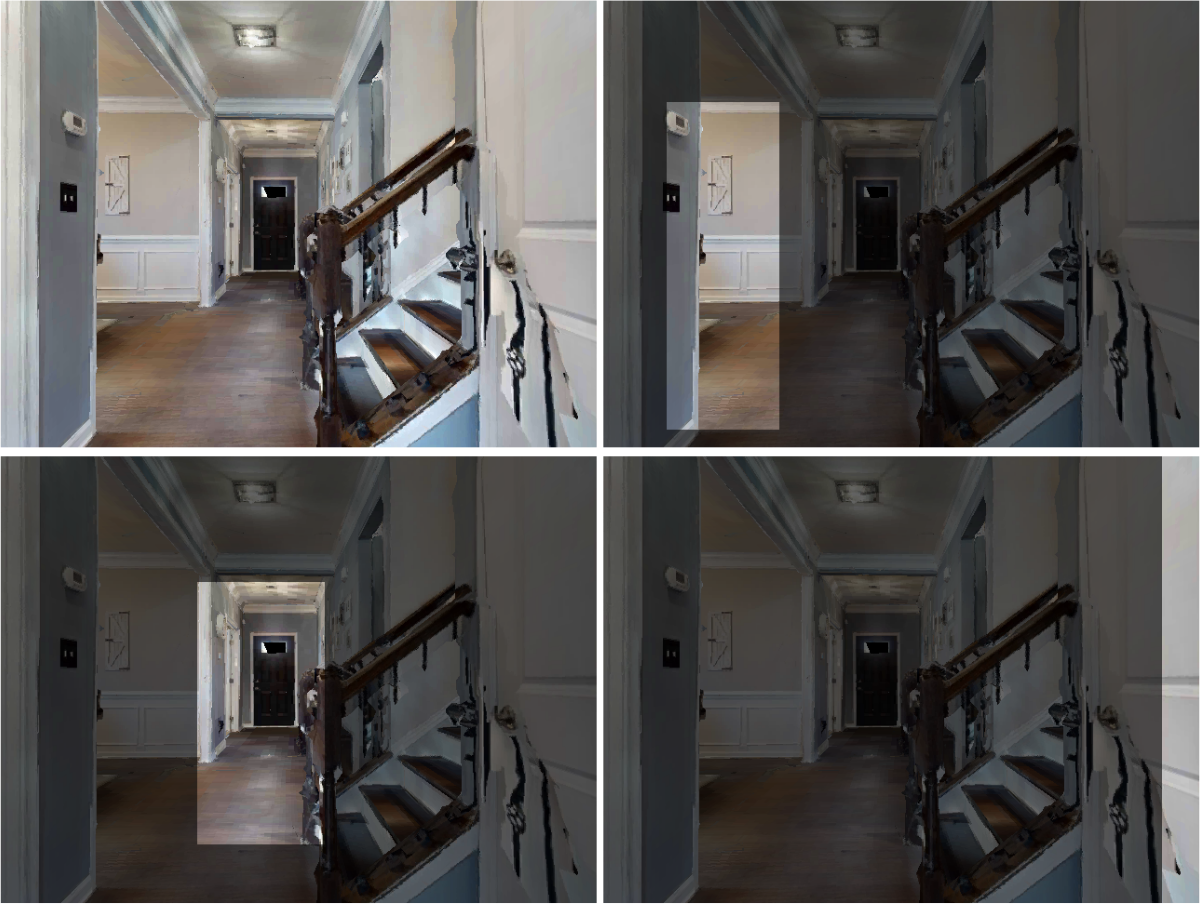}
    \end{subfigure}
    \hfill
    \begin{subfigure}[b]{0.8\linewidth}
         \centering
         \vspace{1.25em}
         \includegraphics[width=\linewidth]{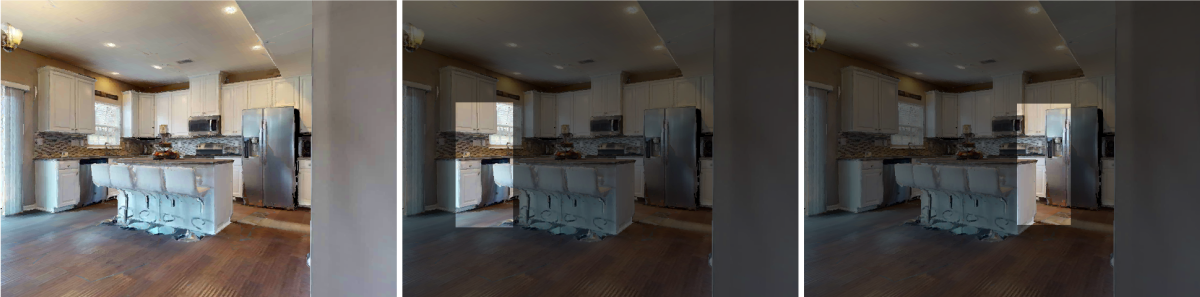}
    \end{subfigure}
    \caption{
    Visualizations of RGB images (top left of each block) and the masks used to compute visual features for each frontier.
    The masks are the highlighted parts of the darker images, localized around each frontier.
    Some frontiers are located on image borders and their mask is computed as a vertical stripe, see the last image in the top blocks.
    }
    \label{supp:frontier-masks}
\end{figure*}

\noindent \textbf{Object merging.}
During the object point cloud merging phase, we match two objects $A$ and $B$, with respective point clouds $P_A$ and $P_B$, by computing the ratio of their points at a distance smaller than $\delta=3cm$ from the other point cloud.
We merge $A$ and $B$ if at least one of these two ratios is above $50\%$.

\noindent{\textbf{Frontier mask for visual features.}}
We show examples of RGB images and the masks applied to these images to extract frontier visual features in~\Cref{supp:frontier-masks}.
These localized masks are computed following the procedure described 
in Section 3.3 of the main paper
and focus on specific regions of the image while the Dino-v2 model extracting their visual features processes the whole image, providing more context about the surroundings of each frontier.
The global feature computed in the ablation of Table 1 from the main paper is instead obtained from the whole RGB image, and thus different frontiers may contain the same visual feature.

\noindent{\textbf{Dealing with holes in Habitat simulator.}}
During online navigation in the MP3D and HM3D environments, we set the camera's minimal depth value to 0 meter. 
This allows us to separate pixels corresponding to holes in the scene meshes, for which the Habitat simulator~\cite{savva2019habitat} renders a depth value of 0, and not back-project them in point cloud creation.
If in a frame we observe both pixels with a valid depth and ones corresponding to mesh holes, we set the depth for these hole pixels to the maximum depth of their neighbouring valid pixels.
This often creates obstacles in our maps at the location of holes and avoids considering inexistent frontiers there.
As several scenes also contain holes in the ground or the ceiling, from which the robot sees objects located at other floors, we additionally filter all the objects located at other floors in our object map.

\subsection{High-level goal prediction model}

\noindent{\textbf{Inputs.}}
The VLM used as the base of our navigation model, \ie, LLaVA v1.6~\cite{liu2023llava,liu2023improvedllava}, processes embeddings of size $D=4096$.
Our added tokens for frontiers, objects and paths are each of size $D$.
Visual features for frontiers and objects are encoded using the large version of Dino-v2 with registers~\cite{oquab2024dinov2,darcet2024registers}.
We use the text encoder from ViT-L-patch14-336 version of CLIP~\cite{radford2021clip}, available through the {\em transformers} library~\cite{wolf2020transformers} to encode categories.
All the distances and spatial coordinates are divided by a factor $\tau=10$ before being fed to the model.

\noindent{\textbf{Training.}}
As in LLaVA~\cite{liu2023llava}, we train our model using an AdamW optimizer~\cite{loshchilov2019adamw} and a cosine learning rate scheduler.
During training, we shuffle input frontier and object embeddings separately before feeding them to the transformer.
Early experiments showed no additional benefit of using RGB images at the current step in our goal prediction model.
As the frontier and object representations already contain visual features, RGB images may bring redundant information while significantly increasing the computational cost of both training and inference.
The current image may also not be informative about the long-term path to follow or the exploration to perform, \eg, seeing a textureless wall or a corridor may not tell anything about where the target is located.

\noindent{\textbf{Online navigation}}
We evaluate the last checkpoint saved in training.
The navigation model is queried after each {\em move forward} action to reduce the risk of alternating between two frontiers and getting stuck.

\subsection{Navigation stopping area}
\noindent After an object has been selected as the next goal and if it is already in the explored part of the scene, we compute a stopping area in which the agent is closer than some threshold to the target.
We obtain a target-free navigability map by substracting three elements to the scene navigability one:

\noindent$\bullet$ The target object 2D map, obtained by projecting its point cloud onto a horizontal plane,

\noindent$\bullet$ The 2D map of obstacles placed above or below the target in the scene,

\noindent$\bullet$ All the obstacles placed on lines between the object's center and the 2D locations from which the agent observed the object during its navigation.

\noindent The distance map to the target is then computed using this target-free obstacle map.
We consider as the valid stopping area the locations closer than 80cm from the target in HM3D and 70cm in MP3D.

\section{Evaluations}

\subsection{Evaluation details}

\noindent{\textbf{Metrics.}}
We notice that computing success for the HM3D v2 dataset using the 3D distance to target view points results in a high number of failed episodes in which the agent correctly reaches the target.
Replacing the 3D distance to the target view points with the 2D one in the horizontal plane solves this issue, as several ground-truth target view points appear to have incorrect heights.
We use it for evaluations on the HM3D v2 benchmark exclusively.

\vspace{0.25em}
\noindent{\textbf{ObjectNav benchmarks.}}
We use the same robot configuration as previous work~\cite{chen2023rim,yokoyama2024vlfm} for these benchmarks, \ie, images of size $640\times480$ acquired at $88$cm from the ground with a field of view of $79$°.
The OVON benchmark evolves in the same houses as HM3D but considers a larger set of categories, split into 280 categories seen during training, and 50 synonyms and 49 unseen categories for validation.
We adopt its robot configuration for these evaluations, \ie, images of size $360\times640$ acquired at 1.31m from the ground with a horizontal field of view of $42$°.

\begin{figure*}
    \centering
    \includegraphics[width=0.775\linewidth]{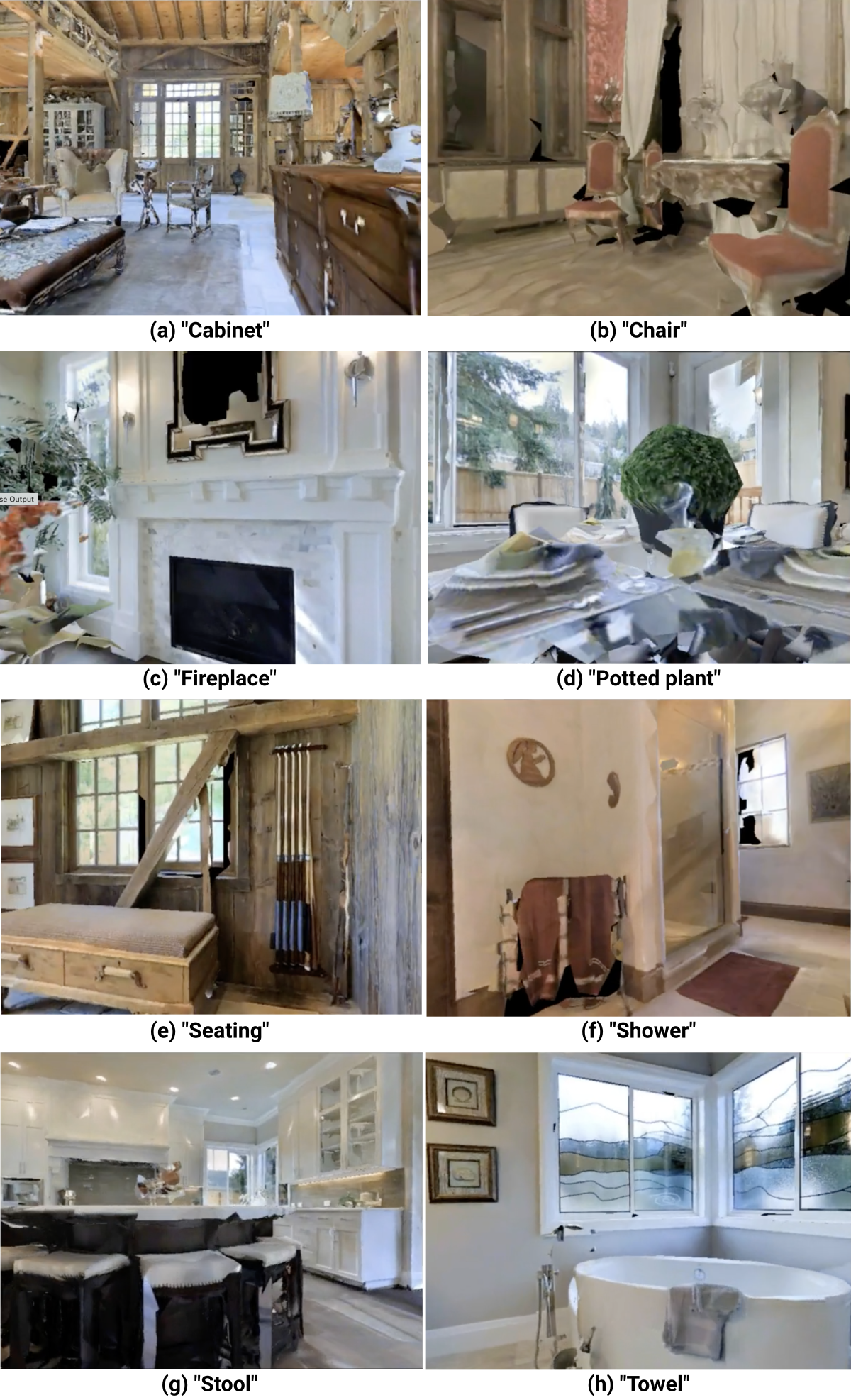}
    \vspace{-0.5em}
    \label{fig:mp3d_annotations}
    \caption{
    Examples of wrong ground-truth annotations in the MP3D dataset.
    The agent decides to stop in these frames with the target category indicated under each image, but the episode is considered a failure as these objects are not part of the GT.
    }
\end{figure*}

\vspace{0.25em}
\noindent{\textbf{Incorrect annotations.}}
The MP3D~\cite{chang2017mp3d} and HM3D v1~\cite{ramakrishnan2021hm3d,habitatchallenge2022} datasets contain incomplete ground-truth annotations, \ie, some correct objects are not annotated as goals in several environments or are wrongly considered as part of another category.
We show such examples in~\Cref{fig:mp3d_annotations}.
This impacts the performance of all the methods in the MP3D benchmark, partially explaining the performance gap compared to HM3D.
Annotations in the HM3D v2~\cite{habitatchallenge2023} are of a higher quality, resulting in higher SR and SPL on the v2 split than the v1.

\vspace{0.25em}
\noindent{\textbf{Computation and memory costs.}}
FOM-Nav uses 20Gb of GPU memory: 15 for LLaVA and 5 for Mask2Former, Dino-v2 and the map storage.
Our maps are updated in 130 ms, with most time spent on point cloud cleaning using DBSCAN.
A forward pass through LLaVA takes 100 ms, using FlashAttention-2~\cite{dao2024flashattention2}.
Its typical inputs contain 5 frontiers, 100 objects, 50 path tokens and the text ones, resulting in around 200 tokens, which is 2 times smaller than, e.g., Uni-Navid~\cite{zhang2024uninavid}.

\subsection{Additional results}

\begin{figure*}
    \centering
    \includegraphics[width=\linewidth]{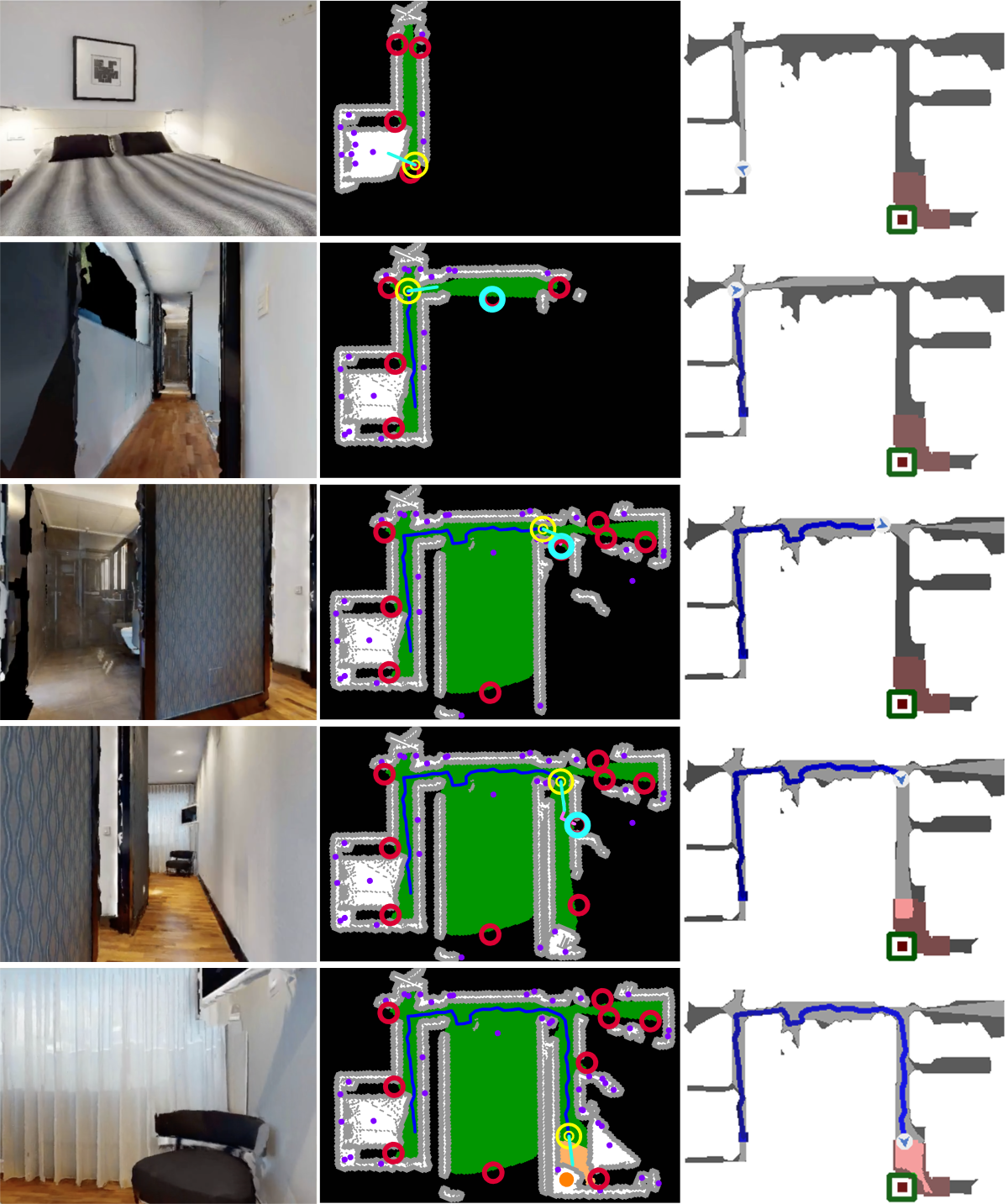}
    \caption{
    Visualization of an episode with navigation to a {\em chair} in an HM3D scene.
    (Left) RGB image seen by the agent.
    (Right) Visualization of the agent's trajectory and the groundtruth stopping area on HM3D GT maps.
    (Center) Our navigation maps built online. 
    Grey pixels correspond to inflated obstacles, the green area is the explored surface, purple dots and red circles are objects and frontiers, respectively, and filled orange and cyan circles are selected next object or frontier.
    The planned and actual paths are shown as pink and navy curves, respectively.
    (Right) Visualization, over the GT map, of the agent's trajectory (navy curve), ground-truth object (white box), and ground-truth stopping area (pink area).
    }
    \label{fig:traj-chair}
\end{figure*}

\begin{figure*}
    \centering
    \includegraphics[width=0.775\linewidth]{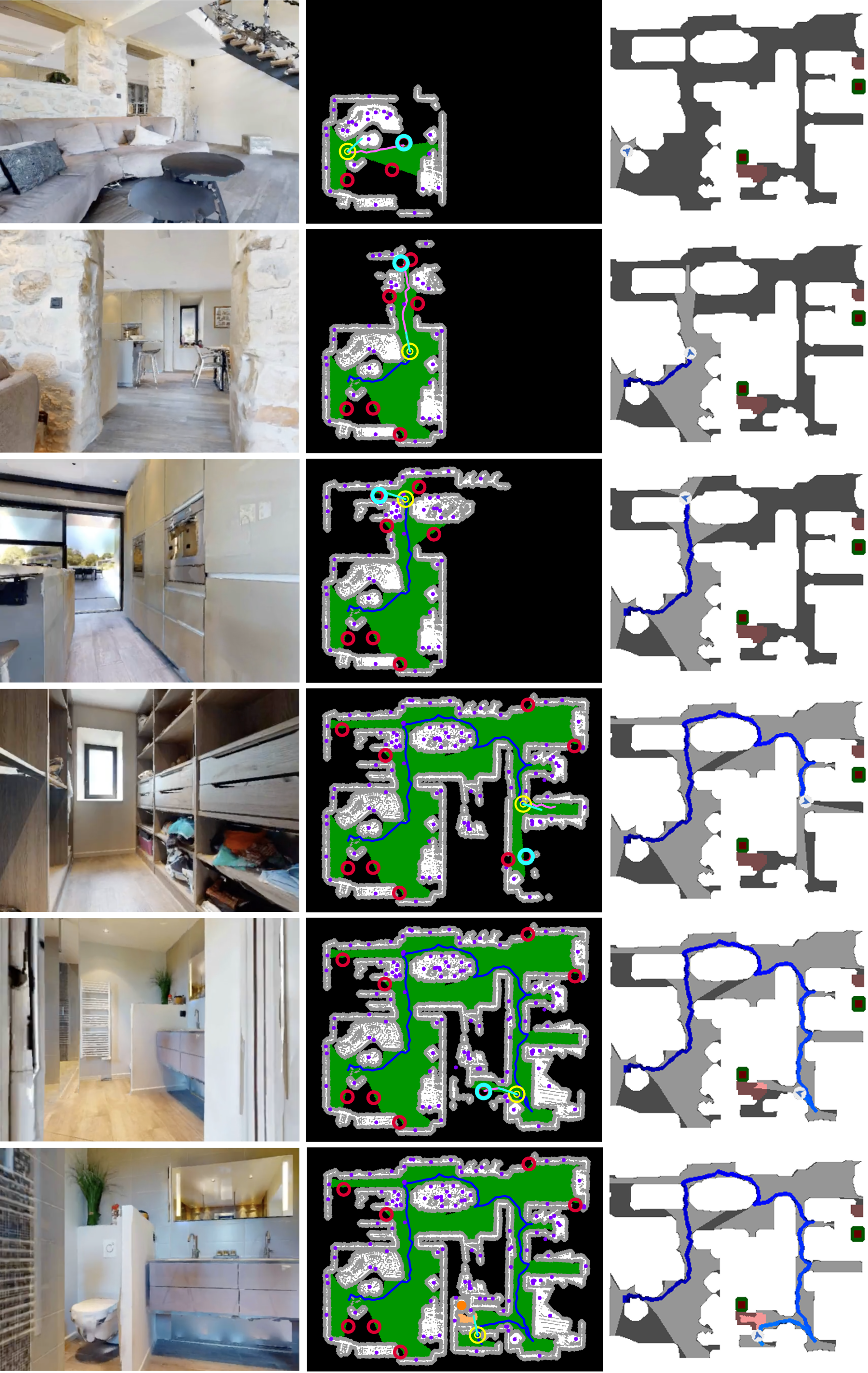}
    \vspace{-0.5em}
    \caption{
    Visualization of an episode with navigation to a {\em toilet} in an HM3D scene.
    (Left) RGB image seen by the agent.
    (Center) Our navigation maps built online. 
    Grey pixels correspond to inflated obstacles, the green area is the explored surface, purple dots and red circles are objects and frontiers, respectively, and filled orange and cyan circles are selected next object or frontier.
    The planned and actual paths are shown as pink and navy curves, respectively.
    (Right) Visualization over the GT map of the agent's trajectory (navy curve), ground-truth object (white box), and ground-truth stopping area (pink area).
    }
    \label{fig:traj-toilet}
\end{figure*}

\begin{figure*}
    \centering
    \includegraphics[width=0.70\linewidth]{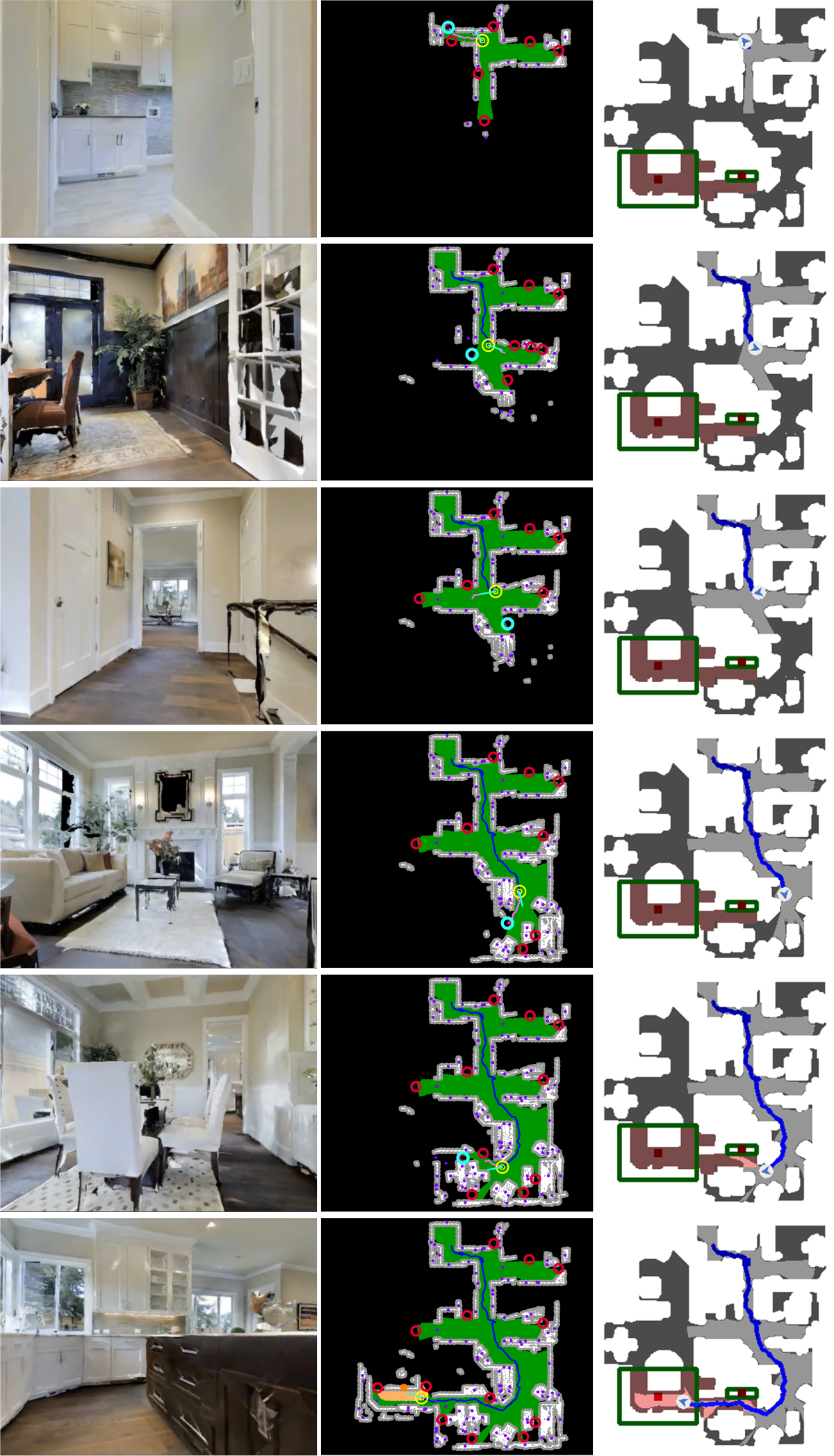}
    \vspace{-0.5em}
    \caption{
    Visualization of an episode with navigation to a {\em counter} in an MP3D scene.
    (Left) RGB image seen by the agent.
    (Center) Our navigation maps built online. 
    Grey pixels correspond to inflated obstacles, the green area is the explored surface, purple dots and red circles are objects and frontiers, respectively, and filled orange and cyan circles are selected next object or frontier.
    The planned and actual paths are shown as pink and navy curves, respectively.
    (Right) Visualization, over the GT map, of the agent's trajectory (navy curve), ground-truth object (green box), and ground-truth stopping area (pink area).
    }
    \label{fig:traj-counter-mp3d}
\end{figure*}

\noindent{\textbf{Navigation visualizations.}}
We show additional examples of successful navigation episodes on the HM3D benchmark in~\Cref{fig:traj-chair,fig:traj-toilet} and on MP3D in~\Cref{fig:traj-counter-mp3d}, where the agent must find the categories {\em chair}, {\em toilet} or {\em counter}.

\begin{table}
\centering
\caption{
Ablation of the base language model used in our method on the HM3D v2 validation split.
}
\label{tab:base_llm_ablation}
\begin{tabular}{lcc} \toprule
& SR & SPL \\ \midrule
Llama 3.2 1B Instruct~\cite{dubey2024llama3} & 74.8 & 47.3 \\
LLaVA v1.6 7B~\cite{liu2023improvedllava} & \textbf{75.8} & \textbf{47.9} \\
\bottomrule
\end{tabular}
\vspace{-1.5em}
\end{table}

\vspace{0.25em}
\noindent \textbf{Ablation of the LLM base model.}
In \Cref{tab:base_llm_ablation}, we evaluate, in the HM3D v2 benchmark, the impact of the base LLM, exchanging a LLaVA 7B model for a smaller and more recent Llama 3.2 1B~\cite{dubey2024llama3}.
While the larger model obtains the best SR and SPL, the gap is narrow, which encourages the use of smaller and more efficient models.

\begin{table}
\centering
\caption{
Results of FOM-Nav on OVON validation splits, training from different versions of our dataset. 
`HM-all' uses all HM3D categories and 
`auto-label' replaces GT labels for synonyms and unseen categories with automatic annotations.
`HM-seen' excludes val synonyms and unseen categories. `+Internet imgs' further includes LVIS and ADE20K segmentation images in training.
}
\label{tab:ovon-fomnav}
\begin{tabular}{ccccccc} \hline
\multirow{2}{*}{\begin{tabular}[c]{@{}c@{}}Training \\ categories\end{tabular}} & \multicolumn{2}{c}{Val seen} & \multicolumn{2}{c}{\begin{tabular}[c]{@{}c@{}}Val synonyms\end{tabular}} & \multicolumn{2}{c}{Val unseen} \\ 
 & SR & SPL & SR & SPL & SR & SPL \\ \hline
HM-all & \textbf{42.5} & \textbf{27.8} & \textbf{36.3} & \textbf{23.4} & \textbf{38.1} & \textbf{25.7} \\
HM-all auto-label
& 37.6 & 24.6 & 29.4 & 18.5 & 30.9 & 20.6 \\
HM-seen & 42.0 & 27.4 & 26.1 & 13.8 & 24.7 & 12.9 \\
\begin{tabular}[c]{@{}c@{}}HM-seen + \\ Internet imgs\end{tabular}
& 41.4 & 27.3 & 29.0 & 16.5 & 30.6 & 17.5 \\ \hline
\end{tabular}

\vspace{-1.5em}
\end{table}

\vspace{0.25em}
\noindent{\textbf{Additional OVON experiments.}}
We collect a dataset to train FOM-Nav on a larger set of categories following the procedure described in~\Cref{sec:dataset}.
We consider all the categories annotated in the HM3D training environments~\cite{yadav2023hm3dsemantics}, and collect 610k and 362k steps over 13.1k and 8k episodes for GT or automatic segmentation.
We then train our model based on a Llama 3.2 1B Instruct model for 6 GPU hours, which is 2 to 3 orders of magnitude less than baselines~\cite{zhang2024uninavid,zhu2025mtu3d}.
Beyond training on all the categories (HM-all) or automatically labeling unseen target categories (HM-all auto-label), we evaluate two more approaches in~\Cref{tab:ovon-fomnav}.
When removing episodes navigating to categories from the synonyms or unseen validation splits (row ``HM-seen''), the performance on the val seen split remain similar but the generalization performance to synonyms or unseen categories drops significantly.
As our model changes both input encoders and the output prediction layer in LLaVA, the benefit from the pretrained LLaVA is limited.
We further investigate extending our navigation dataset with out-of-domain segmentation images from ADE20K~\cite{zhou2019ade20k} and LVIS~\cite{gupta2019lvis} which contain instances of some OVON unseen categories.
We pre-process these images and their ground-truth masks by extracting per-object text and visual features as described in~\Cref{subsec:llava-nav}, set object bounding box coordinates to 0 and randomly sample agent-to-object distances and past trajectories for these inputs during training.
Encouragingly, this joint training substantially improves navigation generalization as shown in the last row.
Future work may explore training FOM-Nav with larger-scale vision-language data to enhance the generalization capabilities.

\begin{figure*}[t]
    \centering
    \includegraphics[width=\linewidth]{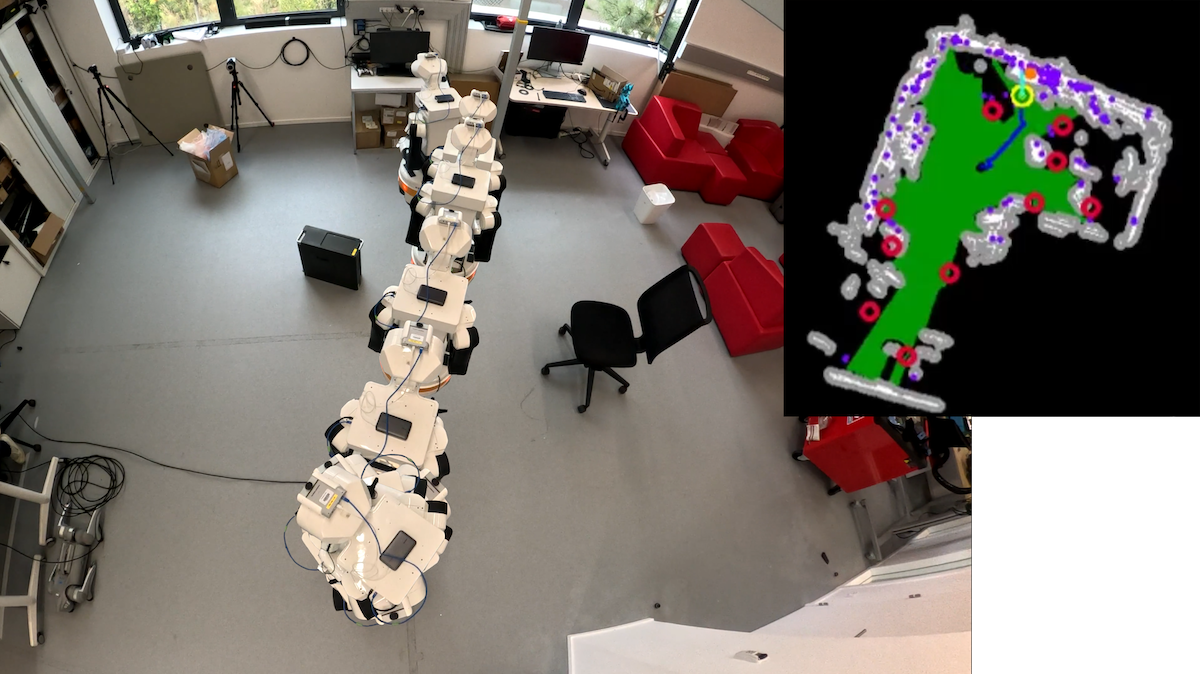}
    \caption{
    Real-world navigation episode, going to a {\em box}.
    (Left) Visualization of the agent's trajectory at different time steps.
    The agent starts by doing a 360° rotation to observe its surroundings and start building a map.
    It then navigates to the best frontier or object as selected by the model, avoiding obstacles on its path, until reaching the target object and stopping there.
    (Right) The constructed top-down 2D map at the end of the trajectory.
    Red circles correspond to frontiers while purple dots are objects.
    The green area represents the explored parts of the scene.
    Obstacles are in white, inflated in grey to indicate the non-navigable area.
    }
    \label{fig:real_world}
\end{figure*}

\vspace{0.25em}
\noindent{\textbf{Real-world deployment.}}
\Cref{fig:real_world} shows an episode during which the robot navigates to a {\em box} in our laboratory, building its map online and avoiding obstacles on its way.

\end{document}